\documentclass{article}

\usepackage[preprint,nonatbib]{neurips_2020}

\usepackage[utf8]{inputenc} % allow utf-8 input
\usepackage[T1]{fontenc}    % use 8-bit T1 fonts
\usepackage{hyperref}       % hyperlinks
\usepackage{url}            % simple URL typesetting
\usepackage{booktabs}       % professional-quality tables
\usepackage{amsfonts}       % blackboard math symbols
\usepackage{nicefrac}       % compact symbols for 1/2, etc.
\usepackage{microtype}      % microtypography

\usepackage{graphicx} %package to manage images
\usepackage{booktabs}
\usepackage{lscape}
\usepackage{longtable}
\graphicspath{ {images/} }
\usepackage[table,xcdraw]{xcolor}
\usepackage{subcaption}

\usepackage{booktabs, makecell, tabularx}
\usepackage{rotating}
\usepackage[export]{adjustbox}

\usepackage{amsmath}

\usepackage{color}

\usepackage{url}

\usepackage{courier}

\usepackage{graphicx,calc}
\newlength\myheight
\newlength\mydepth
\settototalheight\myheight{Xygp}
\newcommand*\inlinegraphics[1]{%
  \settototalheight\myheight{Xygp}%
  \settodepth\mydepth{Xygp}%
   \raisebox{-0.1\mydepth}{\includegraphics[height=0.8\myheight]{#1}}%
}

\title{Using Human Psychophysics to Evaluate Generalization in Scene Text Recognition Models}

\author{%
 \textbf{Sahar Siddiqui} \\
 New York University \\
 \texttt{ss12414@nyu.edu}
 \And
 \textbf{Elena Sizikova} \\
 New York University \\
 \texttt{es5223@nyu.edu} 
\And
 \textbf{Gemma Roig} \\
 Goethe University Frankfurt \\
 \texttt{roig@cs.uni-frankfurt.de} 
\And
 \textbf{Najib J. Majaj} \\
 New York University \\
 \texttt{najib.majaj@nyu.edu} 
 \And
 \textbf{Denis G. Pelli} \\
 New York University \\
 \texttt{denis.pelli@nyu.edu}
}
\begin{document}

\maketitle

\begin{abstract}

  Scene text recognition models have advanced greatly in recent years.  Inspired by human reading we characterize two important scene text recognition models by measuring their domains i.e. the range of stimulus images that they can read. The domain specifies the ability of readers to generalize to different word lengths, fonts, and amounts of occlusion. These metrics identify strengths and weaknesses of existing models. Relative to the attention-based (Attn) model, we discover that the connectionist temporal classification (CTC) model is more robust to noise and occlusion, and better at generalizing to different word lengths. Further, we show that in both models, adding noise to training images yields better generalization to occlusion. These results demonstrate the value of testing models till they break, complementing the traditional data science focus on optimizing performance.

  %While seq2seq networks were designed to be robust to "text alignment" this robustness comes with a cost to other forms of generalization (occlusion, fonts, word length).
  
  % This might be due to the need of dedicated alignment in seq2seq models not necessary in CRNN.
\end{abstract}
%Gemma's comments
\section{Introduction}
% goal 
Ever since neural network models became ubiquitous as computational models in natural language, computer vision, and other domains, understanding their behavior and comparing them to human performance has been a cornerstone in understanding their practical applicability. Here we draw from psychophysics, a branch of psychology that uses behavior to characterize perceptual brain mechanisms, to measure the breaking points of neural networks. This complements the traditional data science emphasis on optimizing performance. Knowing the model's breaking points specifies its domain of operation. This is a powerful method of system identification that is widely used in psychophysics. We focus on the task of scene text recognition which is used in instant translation, handwriting recognition, robot navigation, and industrial automation. These applications often demand that scene text recognition models perform like humans.

\begin{figure*}[t!]
\centering
\includegraphics[width=1\linewidth]{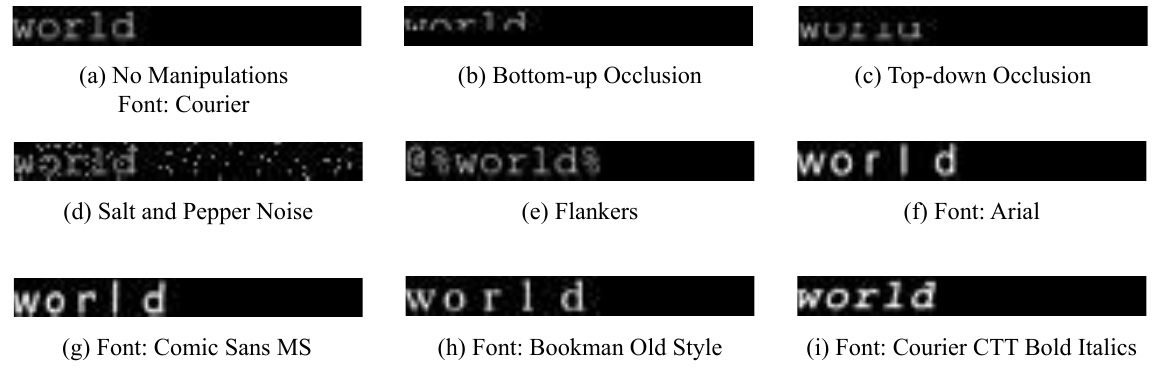} 
\caption{\emph{Sample images of all manipulations.} The stimuli are inspired by the ones typically used in psychophysics human reading experiments \cite{legge2006psychophysics}, which control all the variables, and manipulate the one at study to test generality. }
\label{overview_img} 
\end{figure*}

% challenge
There are two main challenges in scene text recognition. First, scene text recognition is a many-to-many problem, where the input image is characterized using a collection of features, and the output is a sequence of characters. Thus, the resulting loss criterion that the network uses for training needs to address the relative ordering between the inputs and outputs, in addition to any measures of correctness. Second, the input text is typically obtained from natural images, which subjects it to a variety of noises, missing or occluded characters, and other common irregularities (see Fig. \ref{overview_img}). Even after automatic rectification and standardization procedures, the resulting text may exhibit a variety of artifacts that challenge recognition. Additionally, it may not be possible to model all types of these variations in the training set due to their sheer variety. As a result, in measuring how well a network learns to transcribe the seen text, it is important to check its ability to tolerate a variety of corruptions and to see how well it generalizes to unseen perturbations. 

% approach
Two neural network architectures are commonly used for scene text recognition. The first is the convolutional recurrent neural network (CRNN) equipped with a connectionist temporal classification (CTC) loss~\cite{shi2016end}. The second architecture is known as sequence-to-sequence (seq2seq)~\cite{shi2016robust}. The key difference between them is that the seq2seq network uses attention to model alignment between inputs and outputs and models dependencies, whereas the CTC takes the maximum likelihood approach and evaluates all resulting sequences, choosing the likeliest. The two models achieve similar accuracy in standard datasets~\cite{baek2019wrong}.  Yet, analysis of their ability to generalize in different scenarios would shed light on their strengths and weaknesses, to assess suitability for particular applications, and guide improvements.

% overview 
To evaluate and compare the two models, we apply the practice \textemdash widely used in psychophysics \textemdash of testing models till they break. We study how well both networks generalize to noise, occlusion, novel fonts, and different length sequences. These are all dimensions to which human readers generalize readily \cite{huey1908psychology, legge1985psychophysics, tinker1963legibility}. We also study the mistakes that both models make in recognizing letters and words. To control the generalization variables, we perform training and evaluation on synthetically generated images rendered with transformations, such as occlusion, flankers, noise, multiple fonts, etc. We evaluate how training on each of these transformations helps generalize over other transformations.

Our analysis shows that the CTC model (CRNN) holds better generalization properties to noise and occlusions, as well as word length. These results suggest that the attention mechanism used in the seq2seq model that we test introduces a trade-off between accuracy in certain conditions and generalization to other properties. We also find that adding noise to the training images makes the models robust to noise as well as different types of text occlusions. 

In summary, we introduce a new way, drawn from psychophysics, to evaluate scene text recognition models. We measure the breaking points of two state-of-the-art models on dimensions that humans generalize easily to. We find that both models fail to generalize outside of their training domain.

\section{Related Work}
% Here, we  briefly introduce other  works that are related to ours, as those also proposed to evaluate neural networks, but for different tasks than scene text recognition. Then, we introduce relevant psychophysical findings in human reading to relate them to the models' behaviour.  

\paragraph{Evaluation of neural network models.}
Deep neural networks typically have far more trainable parameters than the number of training samples, and have the ability to memorize training distributions and even fit random label distributions~\cite{zhang2016understanding}. Mathematical measures for evaluating generalization 
~\cite{dziugaite2017computing,neyshabur2017exploring,neyshabur2015path} commonly study patterns of training and testing error, and several techniques, such as dropout~\cite{srivastava2014dropout}, early stopping~\cite{prechelt1998early}, and many others~\cite{taylor2017improving}, have been developed to improve generalization.  There is also great interest in interpretability, i.e. understanding how a neural network arrives at its predictions (see \cite{guidotti2018survey} for a survey of approaches). In particular, neural network models have been shown to be vulnerable to small perturbations~\cite{goodfellow2014explaining} and show poor generalization to previously unseen image manipulations~\cite{geirhos2018generalisation}.

% \todo{}{TODO: background on Generalization Analysis, Heiko's paper}
% http://papers.nips.cc/paper/7176-exploring-generalization-in-deep-learning.pdf

\paragraph{Relevant psychophysical findings.}
Psychophysics often varies stimulus properties parametrically to reveal the selectivity of visual mechanisms. Even the first measurements of reading speed noted that human's reading speed is robust to many parameters \cite{huey1908psychology}.  Readers cope with partial occlusion, e.g. when the bottom half of words is covered \cite{huey1908psychology,perea2012revisiting}. Reading speed varies only a few percent across popular fonts, serif and sans serif, and a wide range of text size \cite{legge1985psychophysics,paterson1940make,tinker1963legibility}. Adding noise initially has little effect but sufficiently strong noise makes it impossible to read. Reading in added noise has revealed that human reading is letter by letter and allows identification of the visual mechanism \cite{pelli2003remarkable,majaj2003channel}. Inspired by such studies, we use these psychophysical manipulations in evaluating scene text recognition models. Since these models can be trained from scratch, we applied these parametric variations to both the training and testing phases.

% cite: Huey, E. B. (1908). The psychology and pedagogy of reading. New York: McMillan. Republished in 1968. Cambridge: MA: MIT Press.
%cite:Perea, M. Revisiting Huey: On the importance of the upper part of words during reading. Psychon Bull Rev 19, 1148–1153 (2012). https://doi.org/10.3758/s13423-012-0304-0
% cite: Paterson, Donald Gildersleeve, and Miles Albert Tinker. "How to make type readable." (1940).
% cite: Tinker, Miles Albert. Legibility of print. Iowa State University Press. 1963.
% cite: Majaj, N. J., Liang, Y. X., Martelli, M., Berger, T. D., & Pelli, D. G. (2003). Channel for reading [Abstract]. Journal of Vision, 3(9):813, http://journalofvision.org/3/9/813/, doi:10.1167/3.9.813. [Vision Sciences Society, Sarasota, Florida, May 2003]	pdf
%cite:Pelli, D. G., Farell, B., & Moore, D. C. (2003) The remarkable inefficiency of word recognition. Nature, 423, 752-756.
%cite:Legge, G. E., Pelli, D. G., Rubin, G. S., & Schleske, M. M. (1985). Psychophysics of reading—I. Normal vision. Vision research, 25(2), 239-252.

\section{Scene Text Recognition Models}
Scene text recognition (STR), the task of identifying a sequence of characters from an image seen in the wild, is a challenging and well-studied problem in computer vision. It is related to other image-based text extraction problems such as optical character recognition (OCR)~\cite{impedovo1991optical} and handwritten text recognition~\cite{graves2009offline, messina2015segmentation,wu2017improving}. Most recent work in this domain is neural-network based (c.f. \cite{long2018scene}).

To be succesful, STR must be able to predict outputs of different lengths. To tackle this, multi-stage methods, which first detect characters for each segment of the input and then perform classification, have been proposed~\cite{liu2018char,lou2016generative,yao2014strokelets}. This compressing of the image into a fixed set of features is commonly done using a convolutional image encoder network (e.g. VGG~\cite{simonyan2014very}). 
To avoid the inconvenient segmentation and post-processing steps, most recent methods rely on connectionist temporal classification (CTC)~\cite{graves2006connectionist} or the attention mechanism~\cite{bahdanau2014neural}. 

Both classes of method avoid the need to align the input image and the output characters. To address the variable length input and output sequences that are needed to encode text, they minimize the following loss function:
\begin{eqnarray*}
L(\theta) = -\sum_{(I,S)\in T}\log P(S|I; \theta) = -\sum_{(I,S)\in T} \sum^{L}_{l=1} P(S_l|l;I;\theta),
\end{eqnarray*}
in which $I$ is the input image, $S$ is the predicted string of text, $S_l$ is the predicted character at position $l$, from $L$ total number of characters.  $\theta$ are the network parameters, and $P$ denotes probability. In the following, we describe the CTC and attention mechanisms, and their differences.

\textbf{Connectionist Temporal Classification (CTC).}
CTC-based methods~\cite{shi2016end,wang2017gated} encode sequences into a fixed length encoding vector, predict a single encoding character for each time step, and choose the most likely prediction at the end. 
The key feature of CTC is to use an intermediate text representation, which incorporates  additional blank symbols allowing label repetition and padding text to create a fixed length representation~\cite{graves2006connectionist}. The CTC loss assumes that outputs at each time frame are conditionally independent, and it can be efficiently calculated. Let $\pi$ be a fixed length representation of $S$, and  $B(\pi) = S$ be a many-to-one conversion function. Let $H={h_1,h_2,h.._n}$ be the feature representation of image $I$. Then the CTC loss is the sum of the conditional probabilities of all sequences that map to the correct sequence:
\begin{eqnarray*}
L_{CTC}(\theta) = -\log P(S|H) = -\log \sum_{\pi : B(\pi) = S}P(\pi | H)
\end{eqnarray*}
Note that each $\pi_i$ is conditionally independent: $P(\pi|H)=P(\pi_1|H)P(\pi_2|H)\cdots(\pi_n|H)$ where $n=|\pi|$. While this formulation allows for text prediction in cases where no alignment between image features and characters is available, the CTC loss often makes mistakes, especially in longer sequences. 

\textbf{Attention Decoder (Attn).} 
Alternatively, attention-based methods~\cite{cheng2017focusing, shi2016robust,lee2016recursive,borisyuk2018rosetta,liu2016star} learn a character-level language model, and predict each output character by attending only to a certain predicted region of the input. The alternative formulation with attention does not treat characters as independent as CTC does. Instead, at each step, the decoder outputs character prediction $s_t = softmax(Ws_t+b)$, where $W$ and $b$ are trainable parameters, and $s_t$ is the decoder hidden state at step $s_t = LSTM(y_{t-1},c_t,s_{t-1})$. Here, $c_t$ is the context vector defined as: $c_t = \sum_i \alpha_{ti}h_i$. The attention weight $\alpha_{ti}$ can be computed by $\alpha_{ti} = exp(e_{ti})/ \sum_k exp(e_{tk})$, where $e_{ti}= v^T tanh(Ws_{t-1}+Vh_i+b)$, and $v,W,V$ and $b$ are trainable parameters.

\textbf{CTC vs Attn.} The CTC network is more computationally efficient than the Attn network (faster and fewer parameters). The Attn network achieves higher scores on standard benchmarks, according to the literature. Yet, as we find in our experiments, the Attn network is much more prone to overfitting and does not generalize as well as CTC.

\section{Experimental Setting}
Inspired by human psychophysics, we recommend testing models till they break. In this way, we map the domain of generalization of two state-of-the art scene text recognition models. Here, we first introduce a description of the different conditions and image manipulations that we analyse in our framework. Then, we explain our several image sets and evaluation metrics.

\subsection{Generalization Conditions}
A typical input in a scene text recognition task is a cropped text image containing one word. We synthetically generate such word images with certain manipulations to evaluate scene text recognition models on their abilities to generalise across different types of data. Fig. \ref{overview_img}a shows a word image without any manipulations. We use four major types of manipulations - Occlusion, Font, Flankers, and Noise - of white-letter words on a black background. This simple setting is typically used in psychophysics to control variability and study particular variables or generalizations.

\noindent{\bf Occlusion.} Word images are occluded in two different ways - Bottom-up (Fig. \ref{overview_img}b) and Top-down (Fig. \ref{overview_img}c) where varying amounts of the image are blacked out from the bottom half and the top half of the images, respectively. 

\noindent{\bf Font.} We use five different fonts for rendering word images. These fonts broadly belong to two categories - Serif and Sans Serif. By default, all image sets (except the multiple fonts sets) are rendered using Courier font. Fig. \ref{overview_img} shows a sample of each font.

\noindent{\bf Flankers.} To examine the effect of flankers around words, we add three special characters arranged randomly around the word. These three characters are randomly chosen out of the following nine characters: $ !,\#,\$,\%,\&,*,+,@,\hat{}$ . The position of each special character is determined randomly, but is never placed in between the word letters. Fig. \ref{overview_img}e shows one such sample. These special characters are non-alphabet characters and can easily be distinguished from alphabet characters by a human observer.

\noindent{\bf Noise.} Noise is imposed on the word image in the form of Salt and Pepper noise with probability $p=0.15$ ($p$ is the probability of setting a pixel to either black or white). The specific value of $p$ is chosen to make the stimulus noisy yet recognizable by the human eye.  Fig. \ref{overview_img}d shows a sample image.

\subsection{Dataset}
We generate a word list of 22,878 English words gathered from online dictionaries~\cite{AllScrabble, thefreedictionary,wordtips}. These are lower-case English words of lengths varying from 2 to 7. The word list consists of an equal number of 5-, 6-, and 7-letter words and slightly fewer 2-, 3-, and 4-letter words. The word list is split into training and testing sets with a 70:30 split. Images representing the words are then rendered for training and testing the models. A total of eight image sets are created in this way using image manipulations described earlier. A summary of each of these is presented below:

\noindent\textbf{All words:} Consists of word images without any manipulations. \\
\noindent\textbf{Bottom-up:} Consists of word images with varying percent (0,10,20..100\%) of bottom half occluded. \\
\noindent\textbf{Top-down:} Consists of word images with varying percent (0,10,20..100\%) of top half occluded. \\
\noindent\textbf{Flankers:} Consists of images of words surrounded by special characters acting as flankers. \\
\noindent\textbf{3 fonts:} Consists of word images rendered using one of the three fonts -   \inlinegraphics{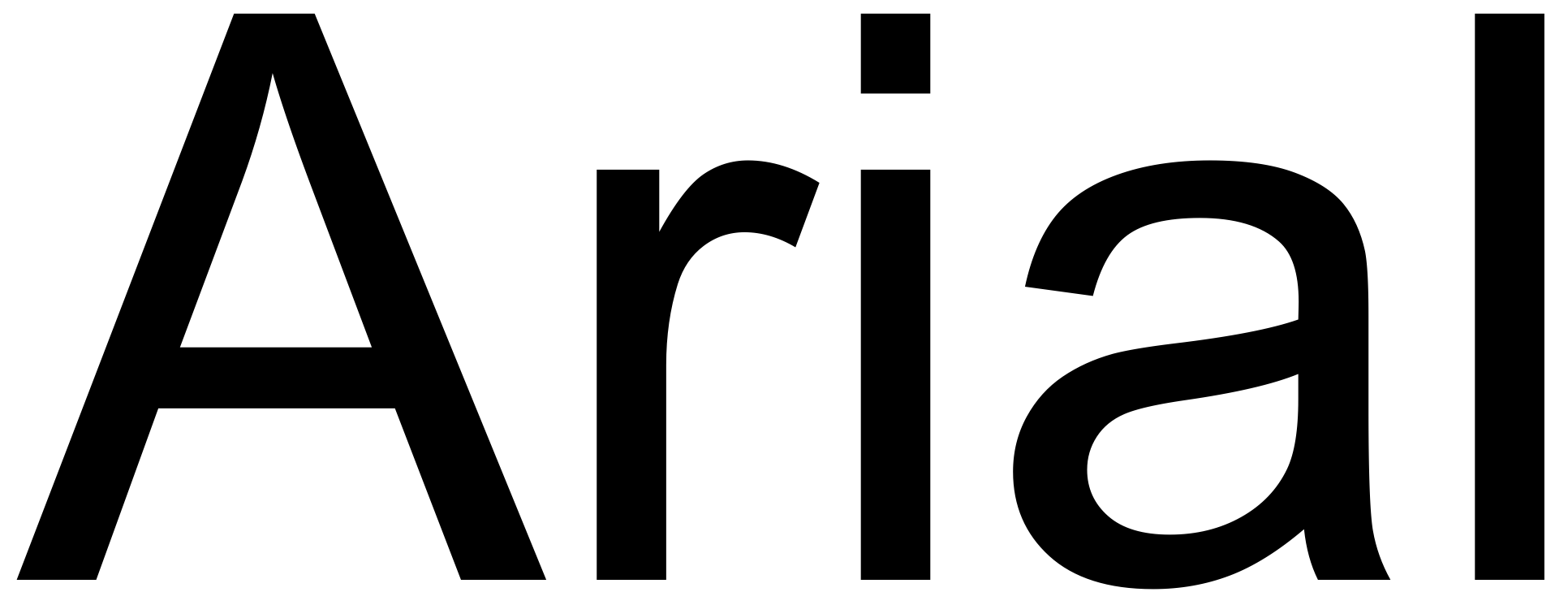}, \inlinegraphics{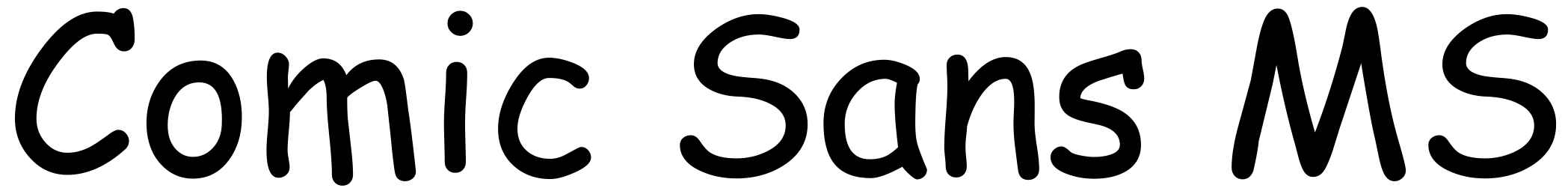},
  \inlinegraphics{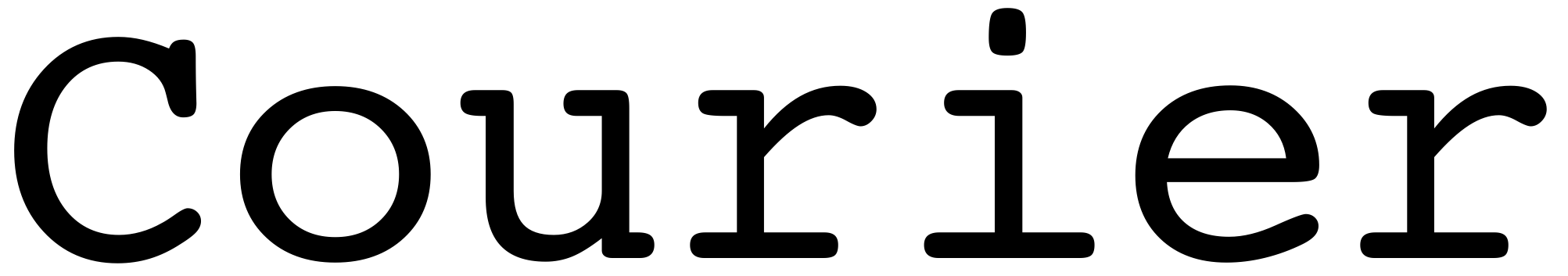}. \\
\noindent\textbf{5 fonts:} Consists of word images rendered using one of the five fonts - \inlinegraphics{courier_img.pdf}, \inlinegraphics{arial_img.pdf}, \inlinegraphics{comicsansms_img.pdf}, {\footnotesize\fontfamily{pbk}\selectfont Bookman Old Style}, and \inlinegraphics{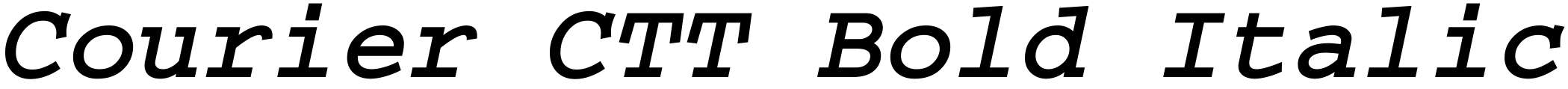}.  \\
\noindent\textbf{Noise:} Consists of word images manipulated using salt and pepper noise. \\
\noindent\textbf{All manipulations:} Consists of word images which each have exactly one of the nine manipulations.\\ 

For faster computation, we pre-render all the images. Each rendered image is a greyscale image of size 100 by 11 pixels. All images are rendered using the Courier font, except the multi-font cases. In addition to the above image sets, experiments are also conducted on three subsets of the All Words images. These subsets, namely "5 letter", "4 and 6 letter", and "3 and 7 letter", consist of words of only those specific lengths in both training and testing sets. For a brief overview, Fig.\ref{overview_img} depicts all considered manipulations of one word image.

\subsection{Implementation Details}
We train one CTC model (CRNN)~\cite{shi2016end} and one Attn model (Seq2Seq)~\cite{shi2016robust} per each of the training sets. Both models are implemented using Pytorch and are trained from scratch. Weights are initialised using the Kaiming He \cite{kaiminghe} initialisation for Attn model and from a normal distribution with mean 0 and variance 0.02 for CTC model. Adam optimizer is applied for training the Attn model and Adadelta optimizer is applied for the CTC model with a batch size of 32 used in both. The images are resized to 100 by 32 pixels for CTC and 280 by 32 for Attn. All the models are trained for up to 100 epochs with an early stopping criteria based on the word-level accuracy of the testing set.

In order to assess the generalization performance of the trained models, we test each model on all the testing sets. The performance results of each trained models are summarised in Table~\ref{tab:heatmaps}. To closely analyze the  trends, we further test the models on smaller subsets of data consisting of varying levels of occlusions, varying word lengths, and multiple fonts. These subsets consist of 100 word images from each subcategory of data manipulations. To account for the uncertainty in data, each of these experiments is performed 5 times, reporting mean accuracy and standard deviation (Fig. \ref{word_length_compare}, \ref{percent_occlusion}, \ref{btt_vs_ttb}, \ref{font_compare}).  

\subsection{Evaluation Metrics}
In a scene text recognition task, given a cropped word image, the model needs to predict the observed string of letters. We evaluate the performance of each model using several evaluation metrics: word-level accuracy, character-level accuracy, word length accuracy, and edit distance. In standard benchmarks, typically the word-level accuracy is used. Using more metrics allows for a more in-depth view of the models' behavior.

\noindent{\bf Word-level accuracy:} This metric evaluates whether all the characters of the predicted word ($P$) and the target word ($T$) match and are in the same position: $A_{\text{word}} = \left( P = =  T\right)$.

% \begin{equation}
%     A_{\text{word}}=\begin{cases}0,\text{ predicted word $\neq$ target word}\\ 1,\text{ predicted word =  target word}\end{cases}
%     % A_{\textit{word accuracy}} = \frac{\textit{Number of correct predictions}}{\textit{Total number of words}}
%     \label{word_accuracy}
% \end{equation}

\noindent{\bf Character-level accuracy:} This metric calculates the number of characters correctly recognized. It measures the number of correctly identified characters, ignoring permutations. This metric gives a soft penalty on partial word matches. We normalize the character accuracy ($C$) by the total number of characters ($N$) in the target word to get the accuracy: $A_{\text{char}}= C/N$.

% \begin{equation}
%     A_{char}= \frac {C} {N}
%     % \sum ^{N}_{i=1}1.\left( predicted_{char\_i} == target_{char\_i}\right)
%     \label{char_accuracy}
% \end{equation}

\noindent{\bf Word Length accuracy:} This metric evaluates whether the predicted word ($P$) has the same length as the target word ($T$): $A_{\text{length}} = \left( \text{length}_{P} = = \text{length}_{T} \right)$. 
% \begin{equation}
%     A_{\text{length}}=
%     \begin{cases}
%         0, \text{length}_{\text{predicted}} \neq \text{length}_{\text{target}} \\ 
%         1, \text{length}_{\text{predicted}} = \text{length}_{\text{target}}
%     \end{cases}
%     \label{word_length_accuracy}
% \end{equation}

\noindent{\bf Edit-distance accuracy:} This metric relies on a string metric called as Levenshtein distance to calculate the difference between two strings. It provides a more accurate measure than the character-level accuracy. Specifically, it measures the minimum number of single-character edits (insertions ($I$), deletions ($D$) or substitutions ($S$)) required to change one word into the other. We normalize the number of edits by target word length ($N$) to assess the error and subtract that from $1$ to get the accuracy: $A_{\text{edit}} = 1 - \left((S+D+I)/N\right)$. 
% \begin{equation}
%     \begin{split}
%         A_{\text{edit}} & = 1 - \frac{S+D+I}{N}.
%     \end{split}
%     \label{edit_distance}
% \end{equation}

\section{Results and Analysis}
In this section, we provide a quantitative and qualitative evaluation of two scene text recognition models on each image set. In Table~\ref{tab:heatmaps}, we present heatmap summaries of the evaluation metrics for each model. It can be observed that at this degree of training, the models are doing fairly well with characters, but poorly with words. (In Table \ref{tab:heatmaps} the heatmaps in the Character Accuracy and Edit Distance Accuracy rows are lighter than those in the Word Accuracy and Word Length Accuracy rows.) As expected, the models perform best at the manipulations they are trained on (diagonals in all the heatmaps). Further, the models generalize best when trained on all manipulations (all the entries in the right-most column are the darkest in each row). The fact that the non-diagonal squares are lighter shows how poorly the models generalize to other manipulations.

\begin{table*}[p!]
    \begin{tabular}{|c|cc|}
        \toprule
        & 
        \hspace{35pt} \textbf{Attn Model (Seq2Seq)} & 
        \hspace{35pt} \textbf{CTC Model (CRNN)}  \\
        \midrule
        \addlinespace[2pt]
         
        \rotatebox[origin=c]{90}{\textbf{Word Accuracy}} & \raisebox{-0.5\height}{\includegraphics[width=0.45\textwidth]{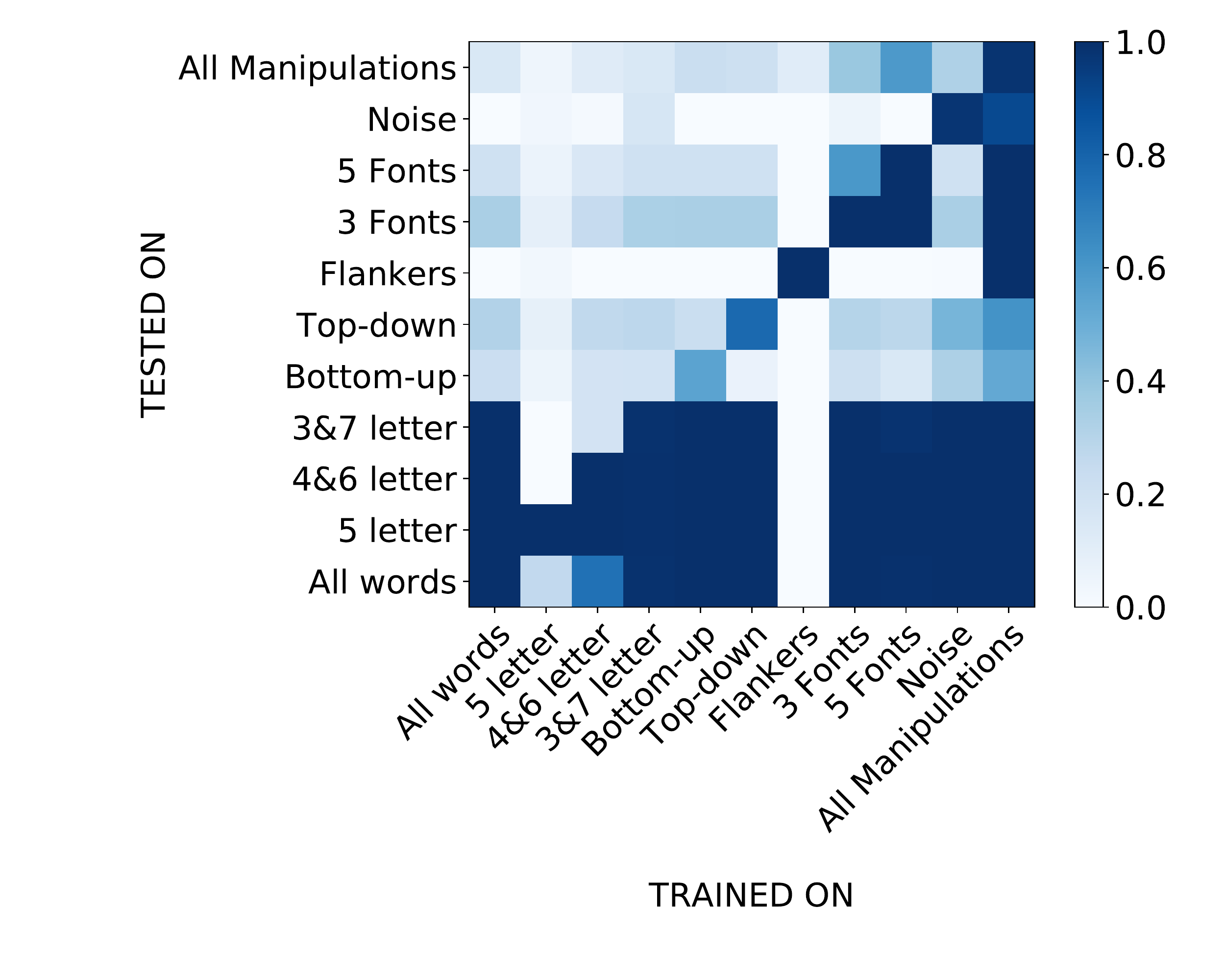}} &
        \raisebox{-0.5\height}{\includegraphics[width=0.45\textwidth]{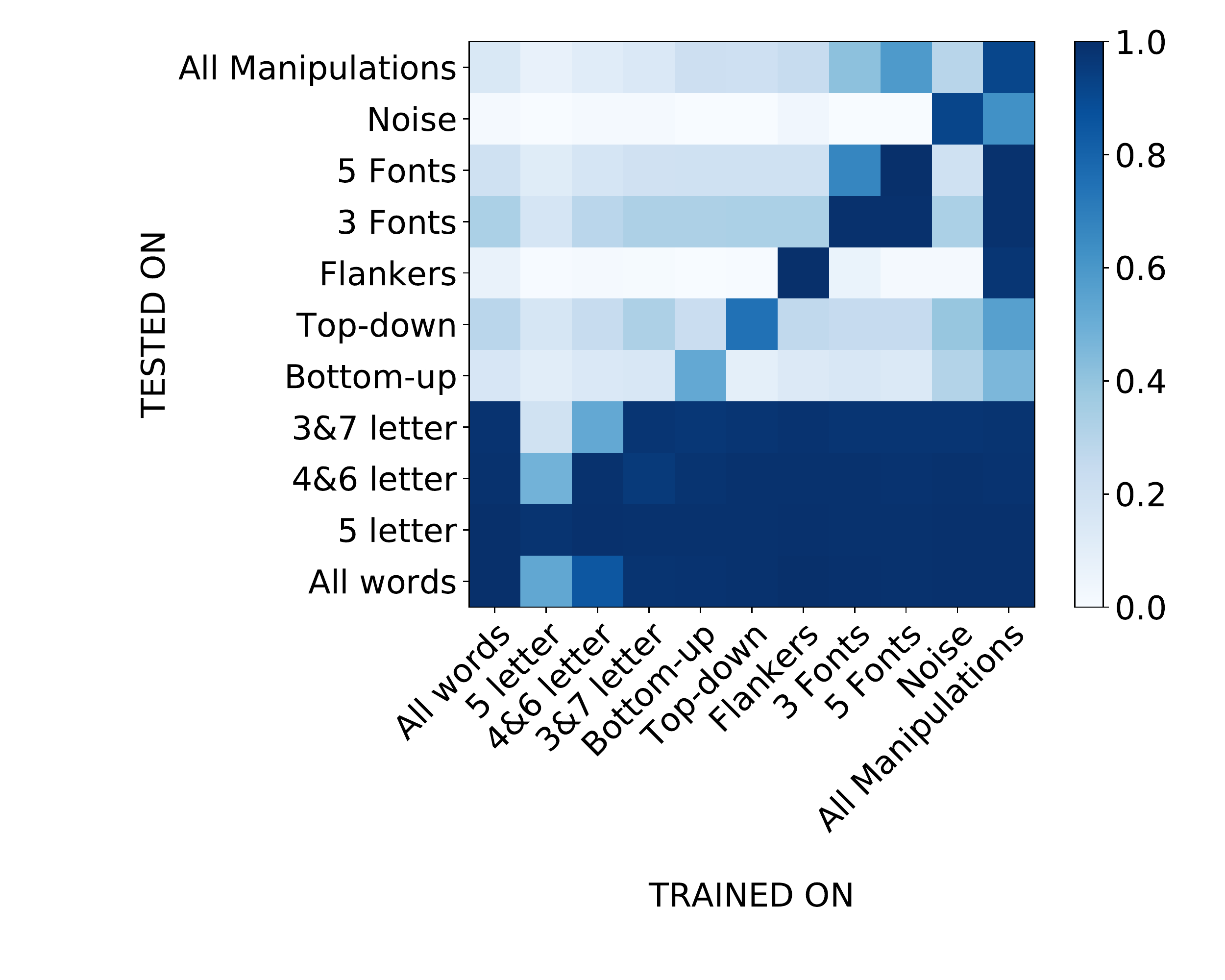}} \\
        
        \toprule
        
        \rotatebox[origin=c]{90}{\textbf{Character Accuracy}} & \raisebox{-0.5\height}{\includegraphics[width=0.45\textwidth]{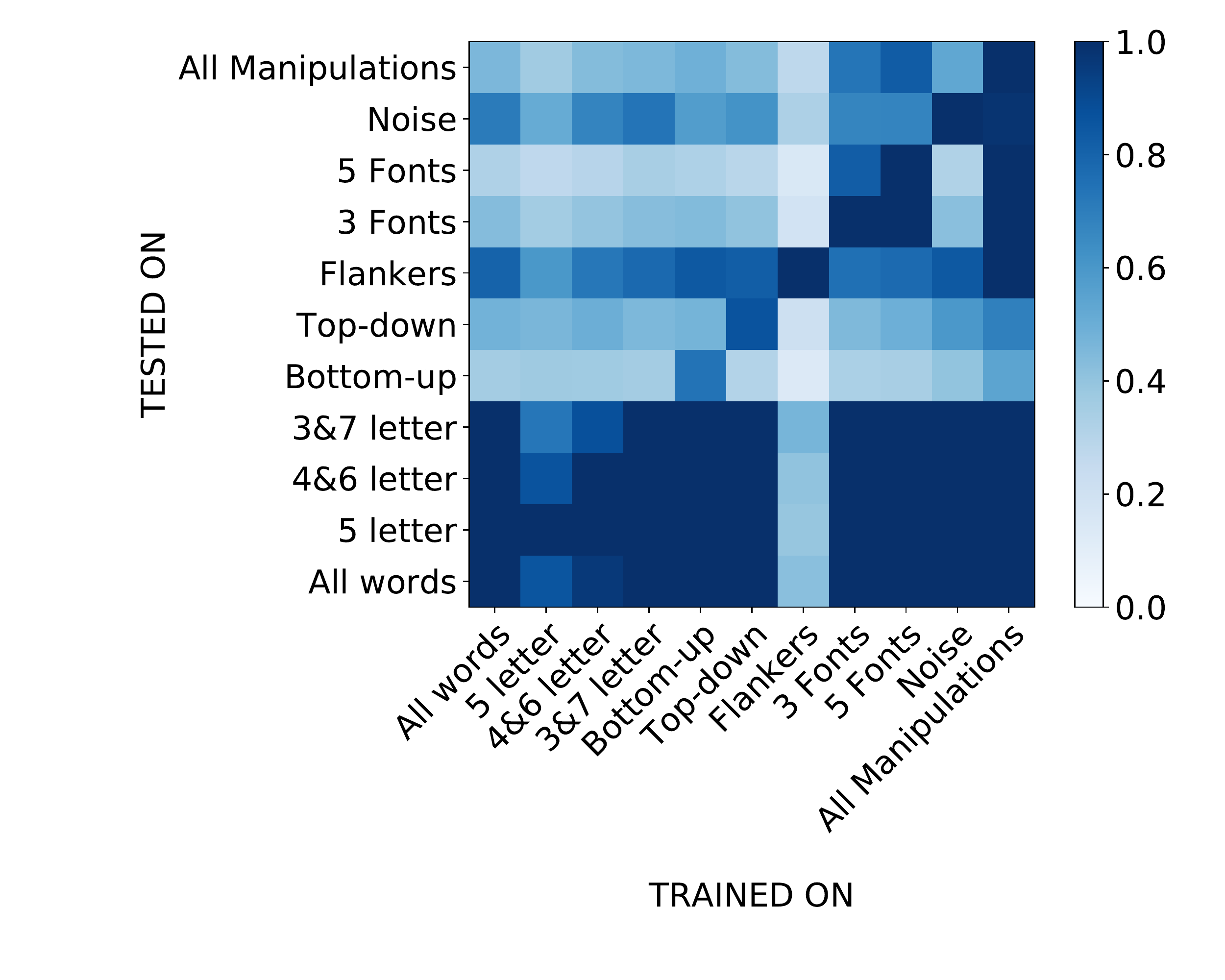}} &
        \raisebox{-0.5\height}{\includegraphics[width=0.45\textwidth]{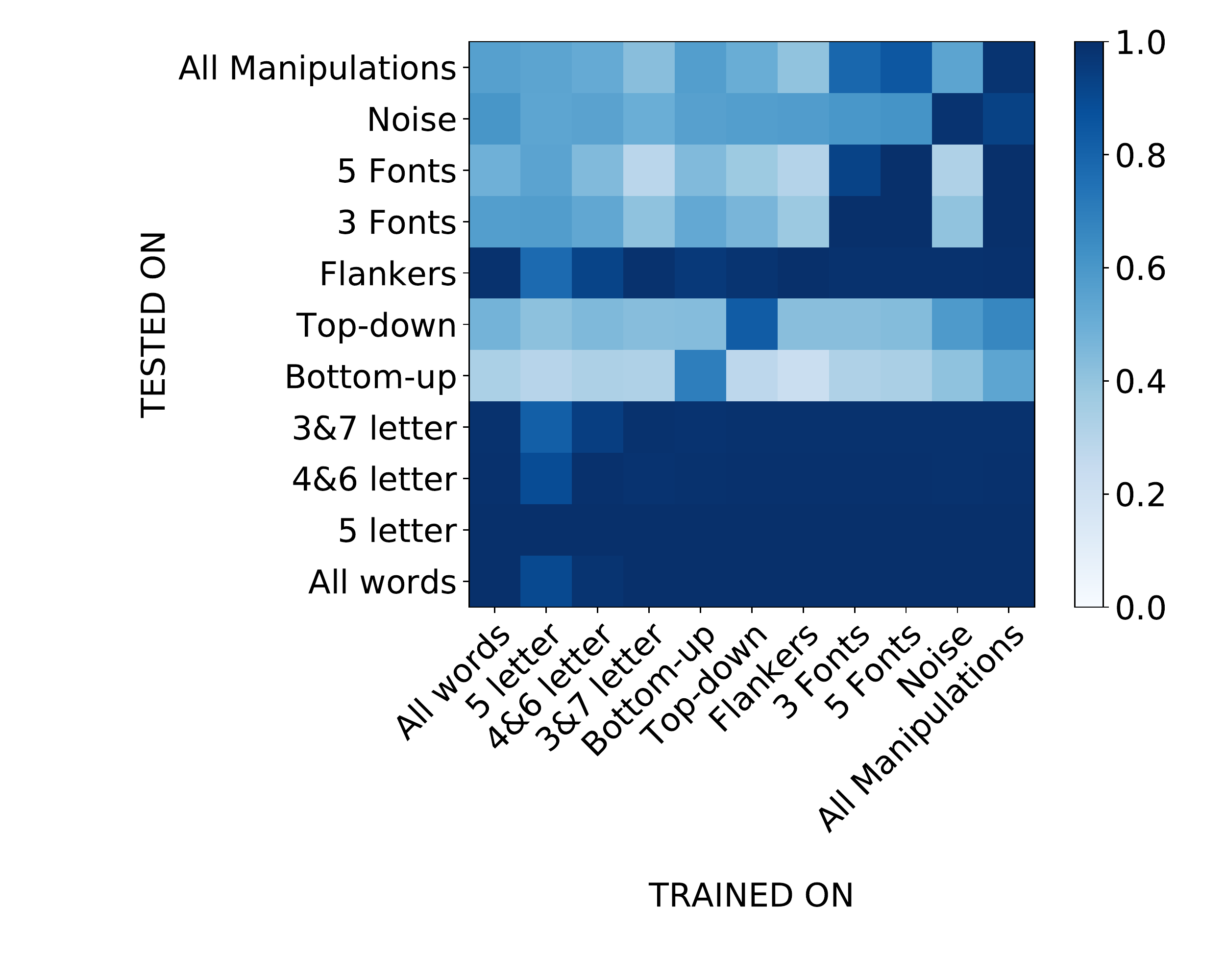}} \\
        
        \toprule
        
        \rotatebox[origin=c]{90}{\textbf{Word Length Accuracy}} & \raisebox{-0.5\height}{\includegraphics[width=0.45\textwidth]{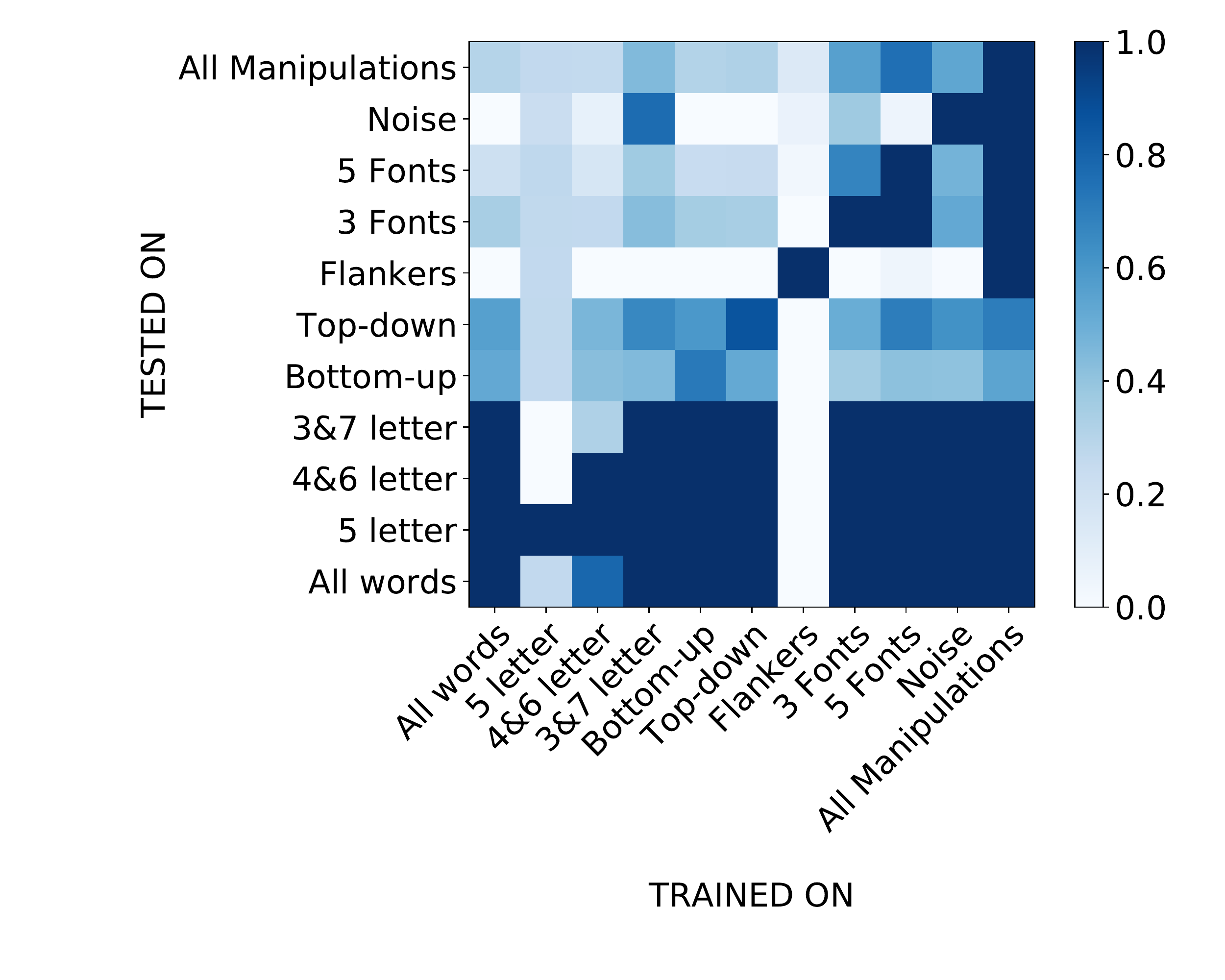}} &
        \raisebox{-0.5\height}{\includegraphics[width=0.45\textwidth]{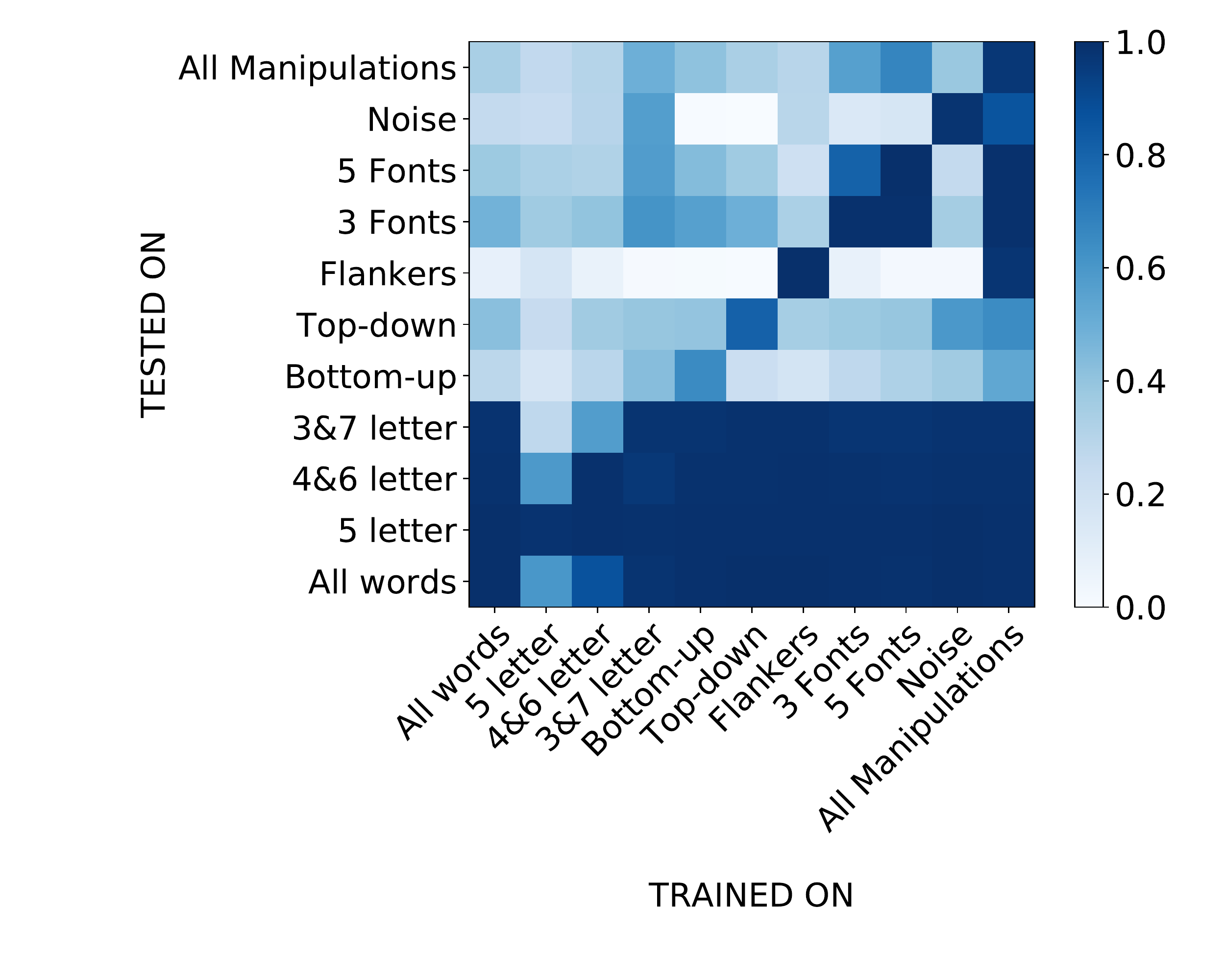}} \\
        
        \toprule
        
        \rotatebox[origin=c]{90}{\textbf{Edit Distance Accuracy}} & \raisebox{-0.5\height}{\includegraphics[width=0.45\textwidth]{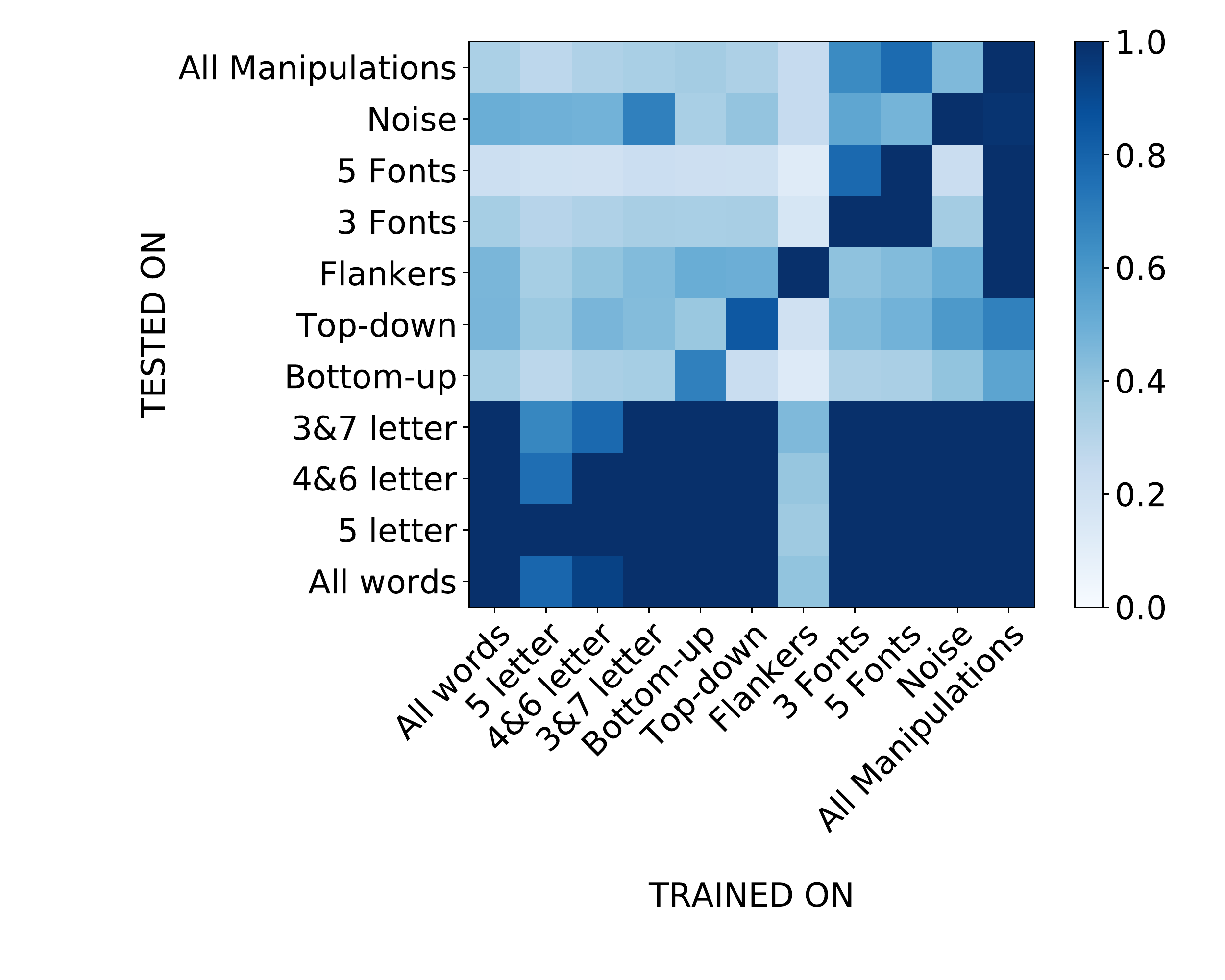}} &
        \raisebox{-0.5\height}{\includegraphics[width=0.45\textwidth]{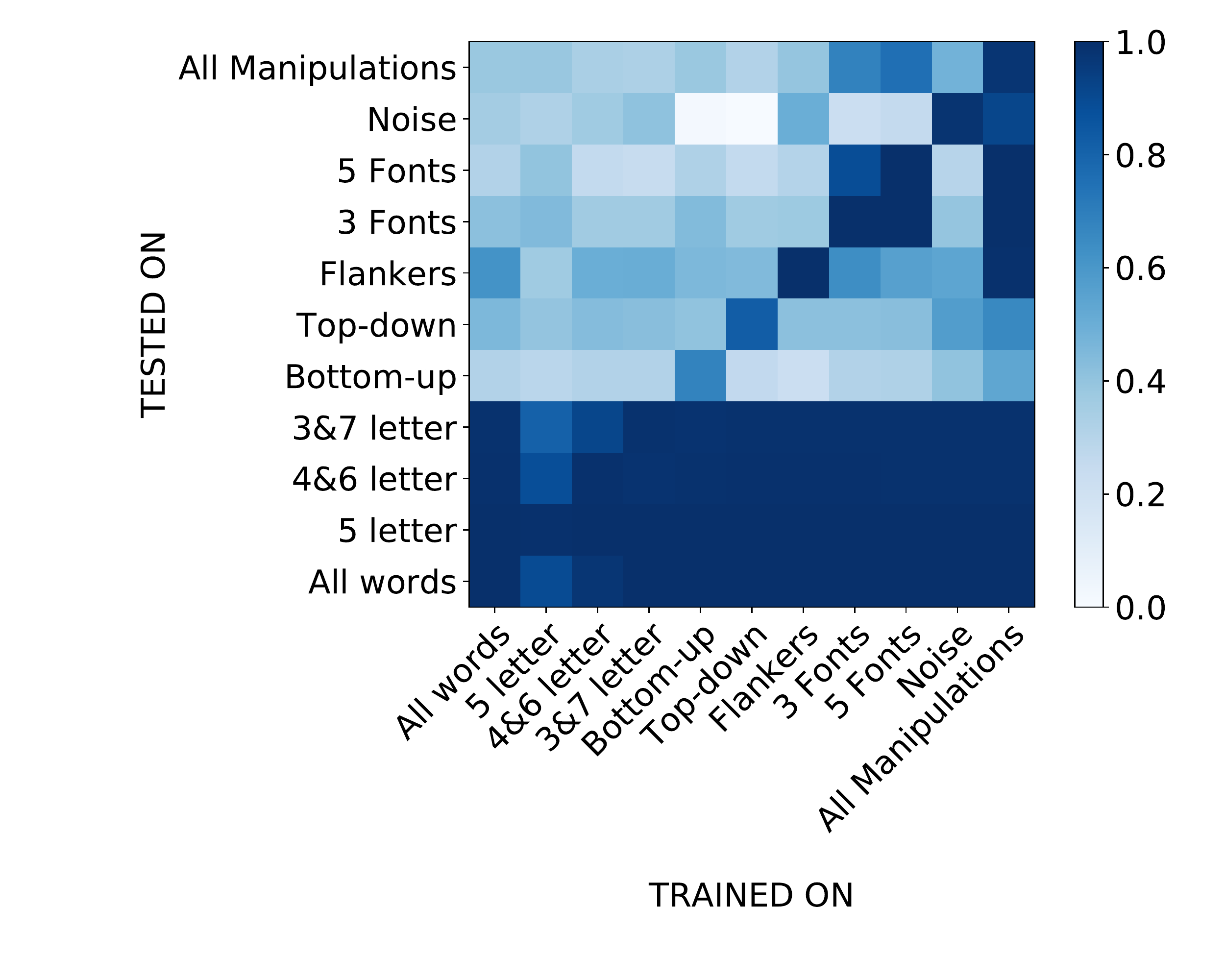}} \\
        
        \toprule
        % \addlinespace[5pt]
        
    \end{tabular}
\centering
\caption{\emph{Performance comparison of Attn and CTC models trained and tested with different types of image manipulations.}}
\label{tab:heatmaps}
\end{table*}

\begin{figure}[t!]
\centering
  \begin{subfigure}[b]{0.4\linewidth}
    \centering
    \includegraphics[width=1\linewidth]{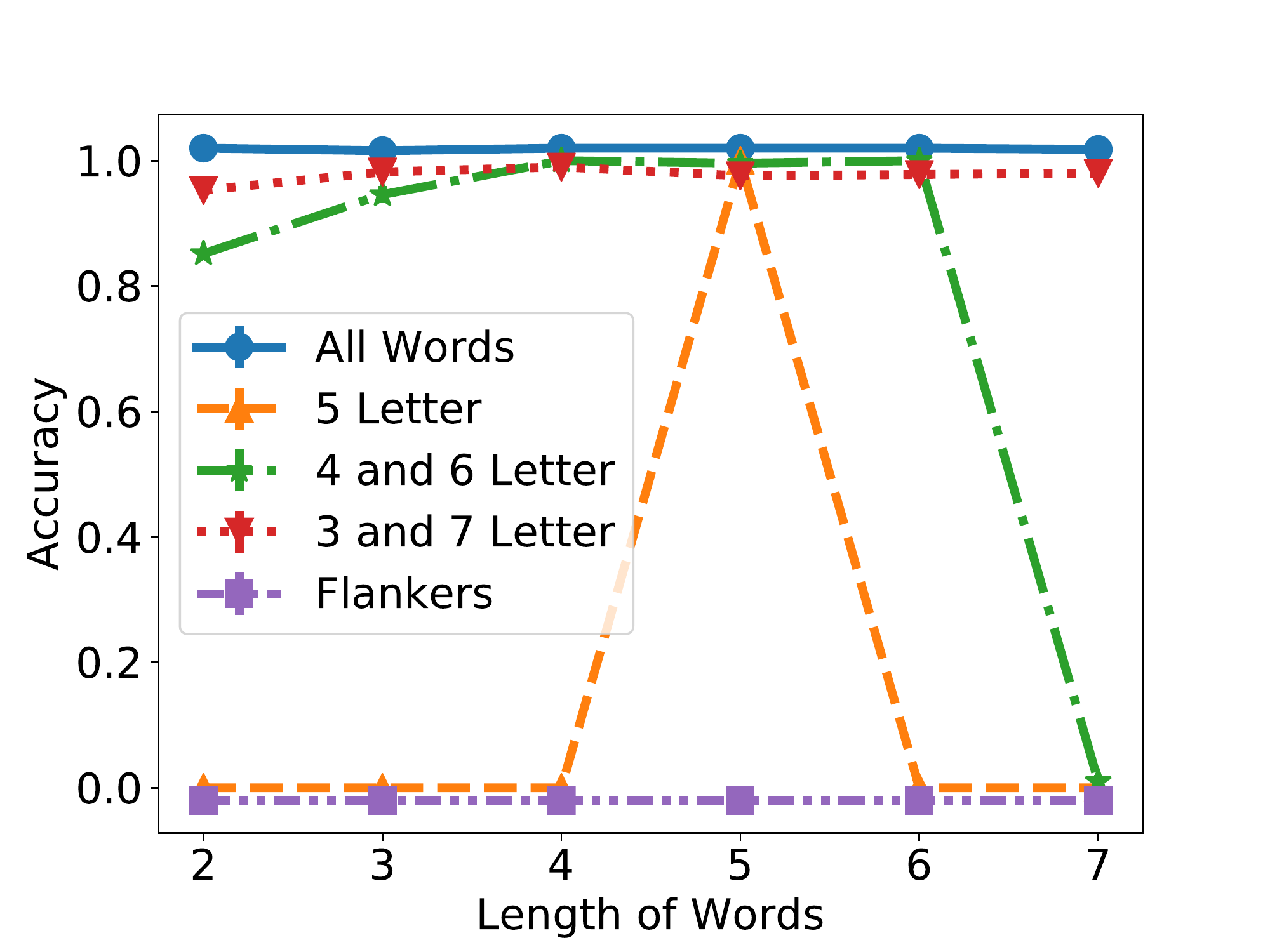} 
    \caption{Attn model (Seq2Seq)} 
    \label{word_length_compare:a} 
    % \vspace{4ex}
  \end{subfigure}%% 
  \begin{subfigure}[b]{0.4\linewidth}
    \centering
    \includegraphics[width=1\linewidth]{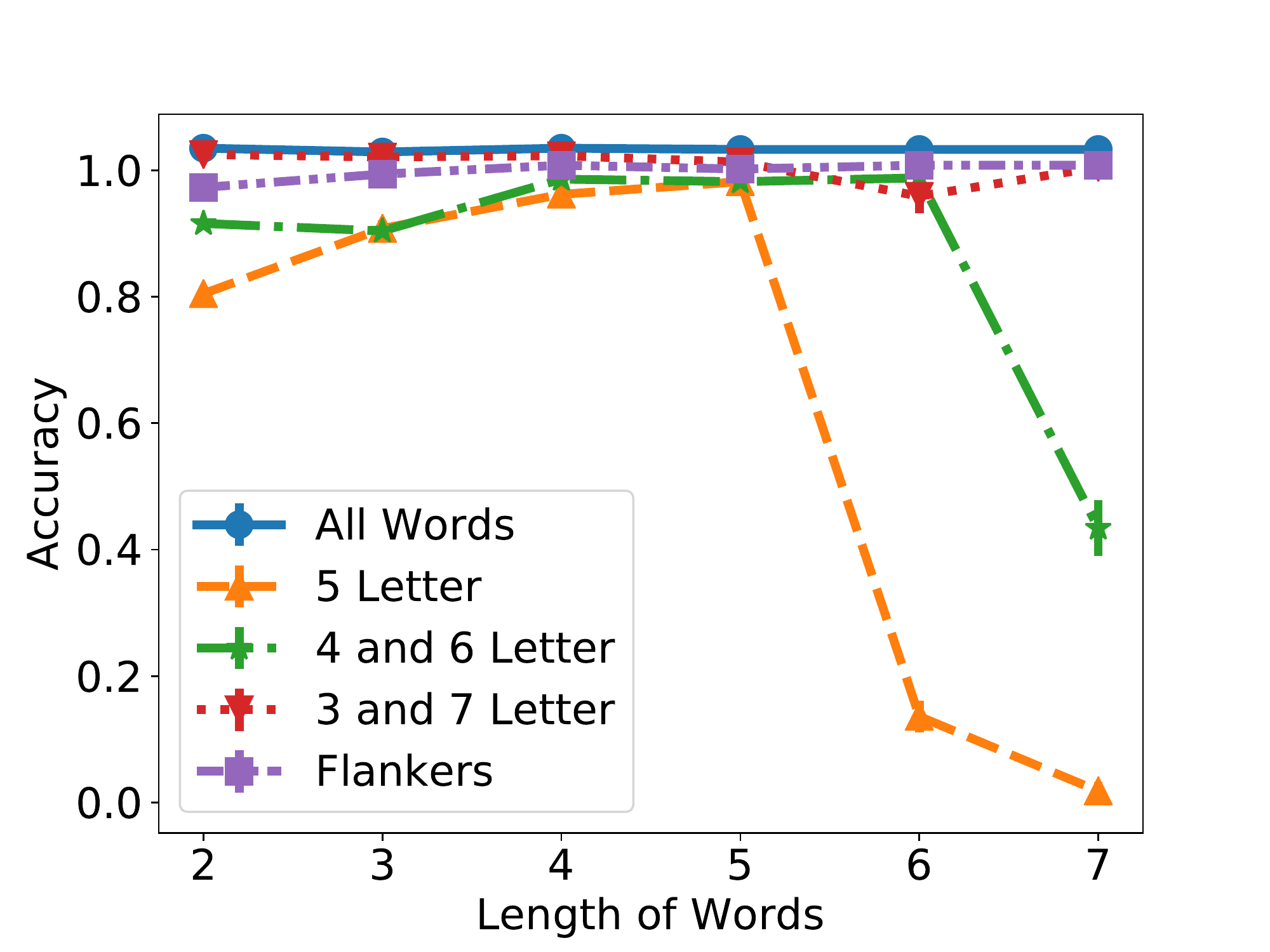} 
    \caption{CTC model (CRNN)} 
    \label{word_length_compare:b} 
    % \vspace{4ex}
  \end{subfigure}
\caption{\emph{Analysis of word accuracy of models evaluated on different word lengths.} Models trained on non-noisy images of all words (blue), only 5-letter words (orange), only 4- and 6-letter words (green), only 3- and 7-letter words (red), and words with flankers (purple) are tested on non-manipulated images of different word lengths. Average accuracy over 5 runs is plotted with error bars indicating +/- standard deviation. Curves have been slightly shifted vertically for clarity.} 
%\todo{}{Curves have been slightly shifted vertically to minimize occlusion. Please put all captions in a sans serif font like Helvetica or Arial.} 
  \label{word_length_compare} 
\end{figure}

\begin{figure}[t!]
\centering
   \begin{subfigure}[b]{0.4\linewidth}
    \centering
    \includegraphics[width=0.7\linewidth]{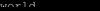} 
    \caption{Sample Bottom-up Occlusion}
  \end{subfigure}%% 
  \begin{subfigure}[b]{0.4\linewidth}
    \centering
    \includegraphics[width=0.7\linewidth]{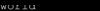} 
    \caption{Sample Top-down Occlusion}
  \end{subfigure}
  
  \begin{subfigure}[b]{0.4\linewidth}
    \centering
    \includegraphics[width=1\linewidth]{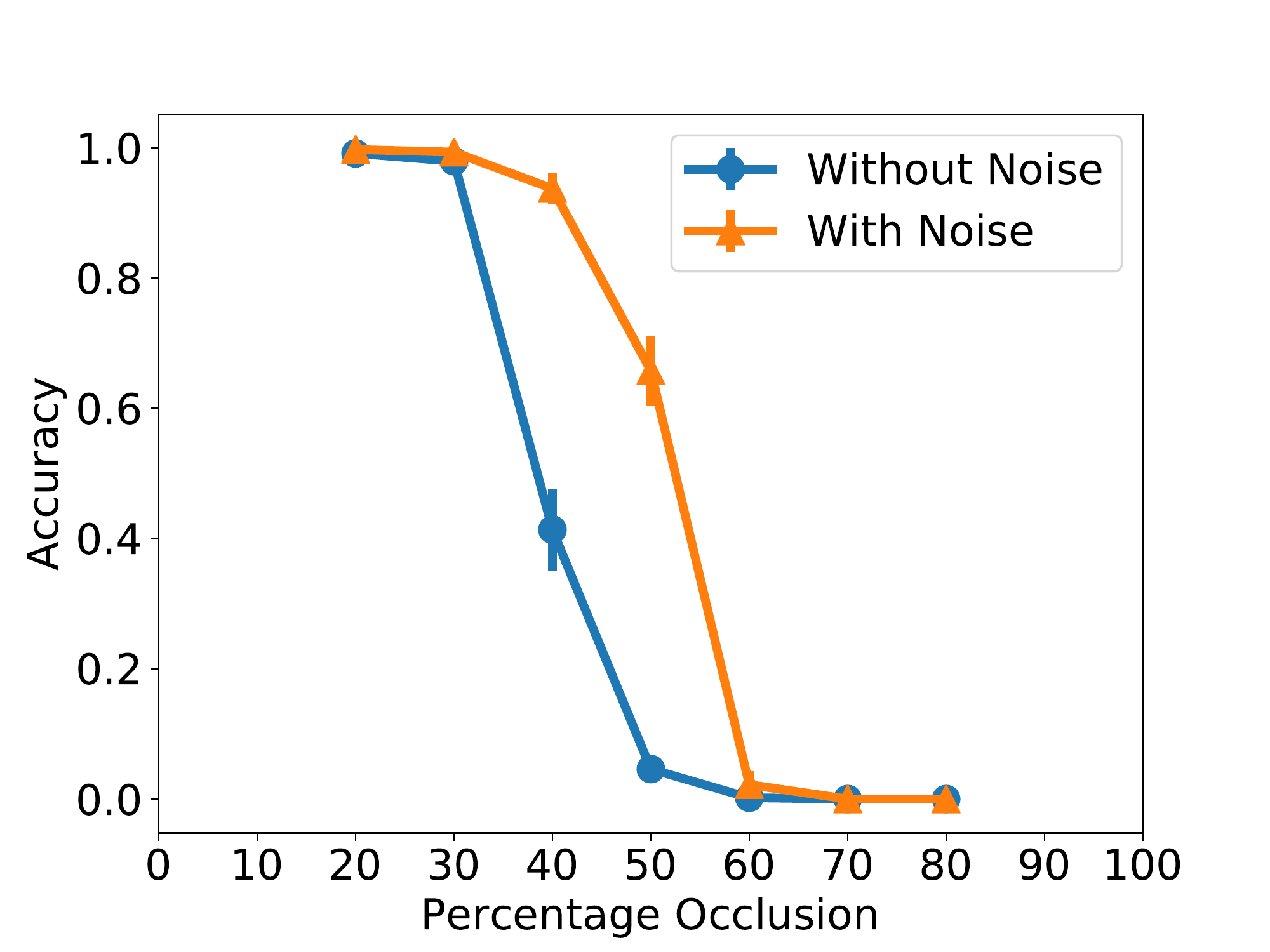} 
    \caption{Attn model: Bottom-up Occlusion} 
    \label{percent_occlusion:a} 
    % \vspace{4ex}
  \end{subfigure}%% 
  \begin{subfigure}[b]{0.4\linewidth}
    \centering
    \includegraphics[width=1\linewidth]{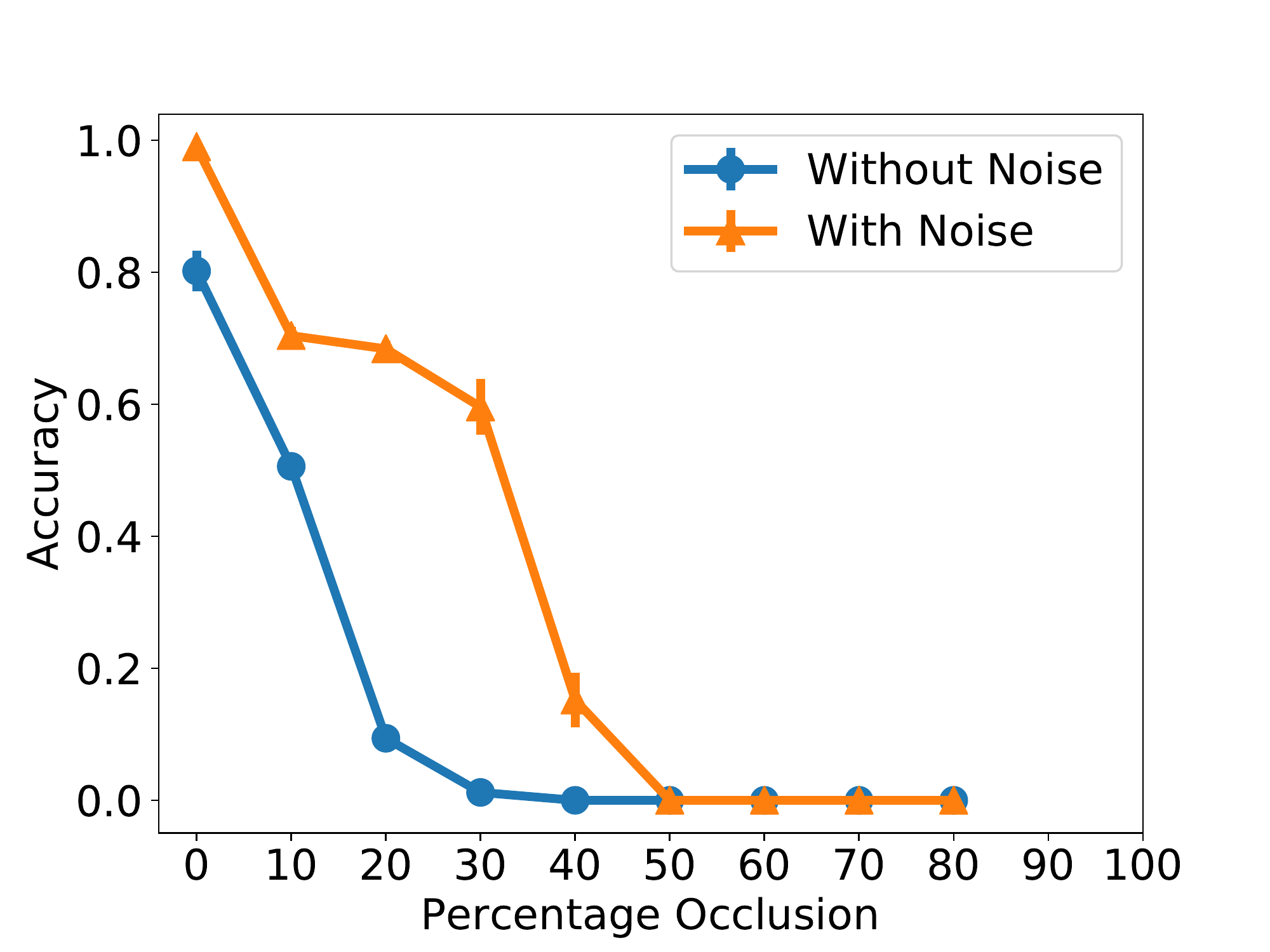} 
    \caption{Attn model: Top-down Occlusion} 
    \label{percent_occlusion:b} 
    % \vspace{4ex}
  \end{subfigure} 
  \begin{subfigure}[b]{0.4\linewidth}
    \centering
    \includegraphics[width=1\linewidth]{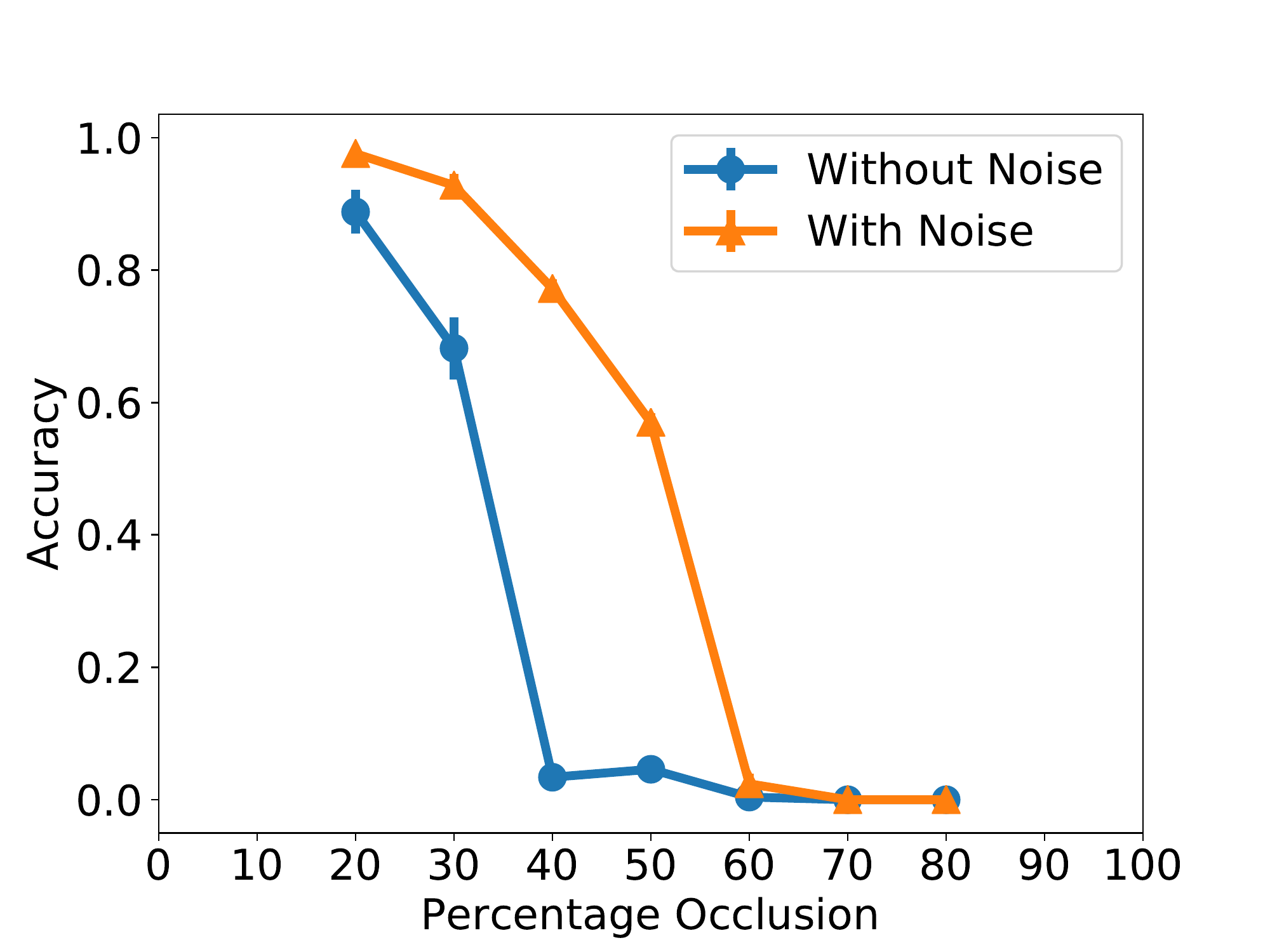} 
    \caption{CTC model: Bottom-up Occlusion} 
    \label{percent_occlusion:c} 
  \end{subfigure}%%
  \begin{subfigure}[b]{0.4\linewidth}
    \centering
    \includegraphics[width=1\linewidth]{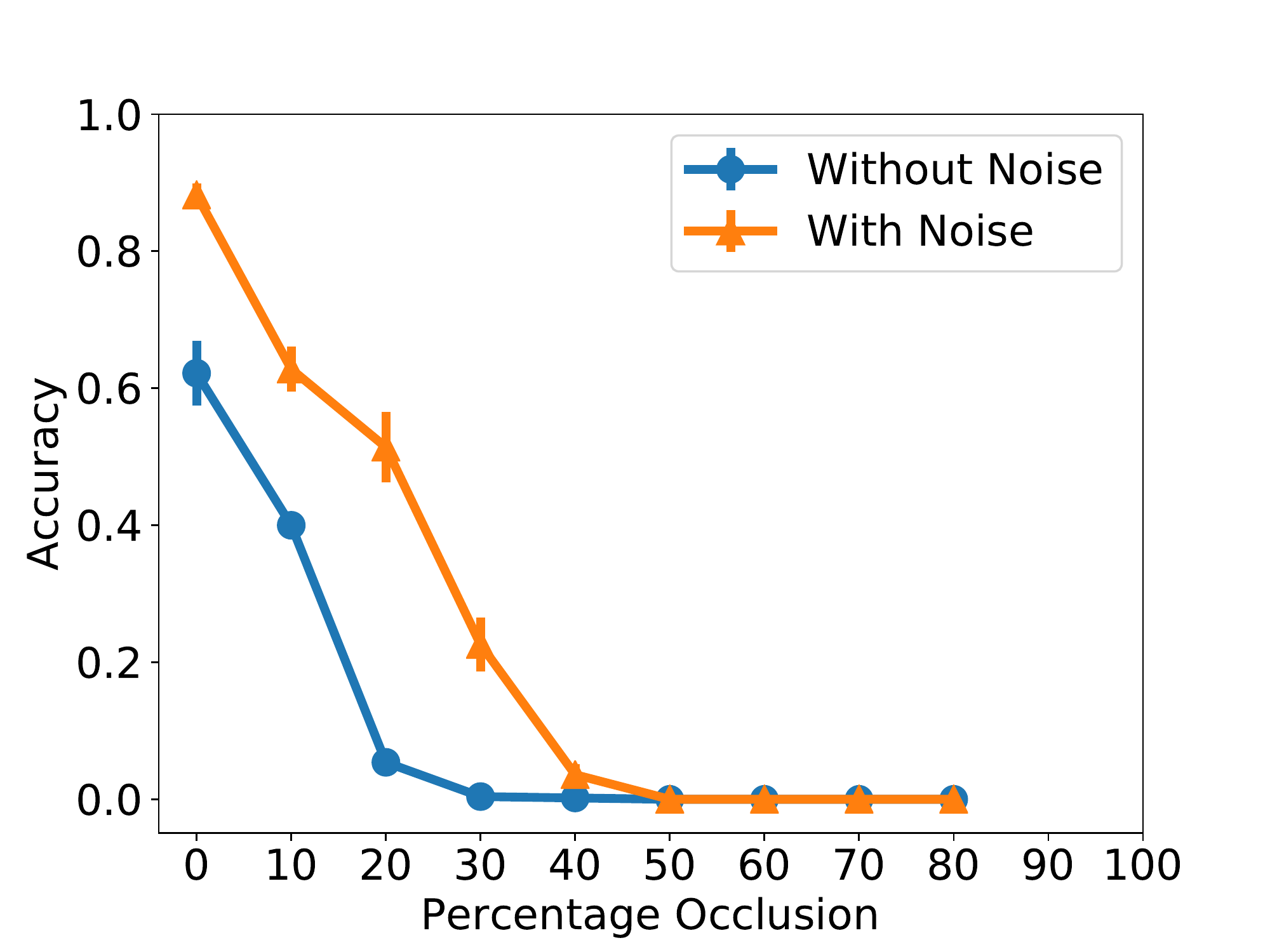} 
    \caption{CTC model: Top-down Occlusion} 
    \label{percent_occlusion:d} 
  \end{subfigure} 
%   \vspace{1ex}
  \caption{\emph{Analysis of word accuracy of models evaluated on various amounts of image occlusion}. Models trained without noise are shown in blue, models trained with noise are shown in orange. Average accuracy over 5 runs are plotted with error bars indicating +/- standard deviation.\vspace*{-0.35cm}}
  \label{percent_occlusion}
\end{figure}

\noindent{\bf Word length.} We train both networks with images of words of specified length to assess how well they generalize to other word lengths. All the heatmaps show that models trained on both 3- and 7-letter words generalize well to all word lengths. However, models trained only on 5-letter words generalize poorly to other lengths.  Furthermore, we observe that the CTC model (CRNN) generalizes to other word lengths much better than the Attn (Seq2Seq) model does. Fig. \ref{word_length_compare} gives a closer look at how each model performs with several word lengths. The Attn model overfits to the trained word length when trained with just 5-letter words or the combination of 4- and 6-letter words.

\noindent{\bf Flankers.}
Fig.~\ref{word_length_compare} reveals that the Attn (Seq2Seq) model trained with flankers cannot recognize words of any length. However, heatmaps for character accuracy and edit distance accuracy (Fig. \ref{tab:heatmaps}) show that both models do identify some characters correctly. A closer look at specific failures reveals that the Attn model attempts to align the input sequence with the output sequence, but does not learn to distinguish alphabetic from non-alphabetic characters. For example, seeing an image of the word ``model", the Attn network trained with flankers predicts ``ode", discarding the surrounding characters as flankers. The CTC model does not suffer from this alignment problem. The CTC model handles flankers perfectly (purple line in Fig. \ref{word_length_compare}).

\noindent{\bf Noise and occlusion.}
A common trend seen in both models is that training on noise helps generalize to occlusion. Fig.~\ref{percent_occlusion} shows that training with noise improves the performance of both models by roughly 18\% in both types of occlusion, bottom-up and top-down.

\begin{figure}[t!]
\centering
  \begin{subfigure}[b]{0.4\linewidth}
    \centering
    \includegraphics[width=1\linewidth]{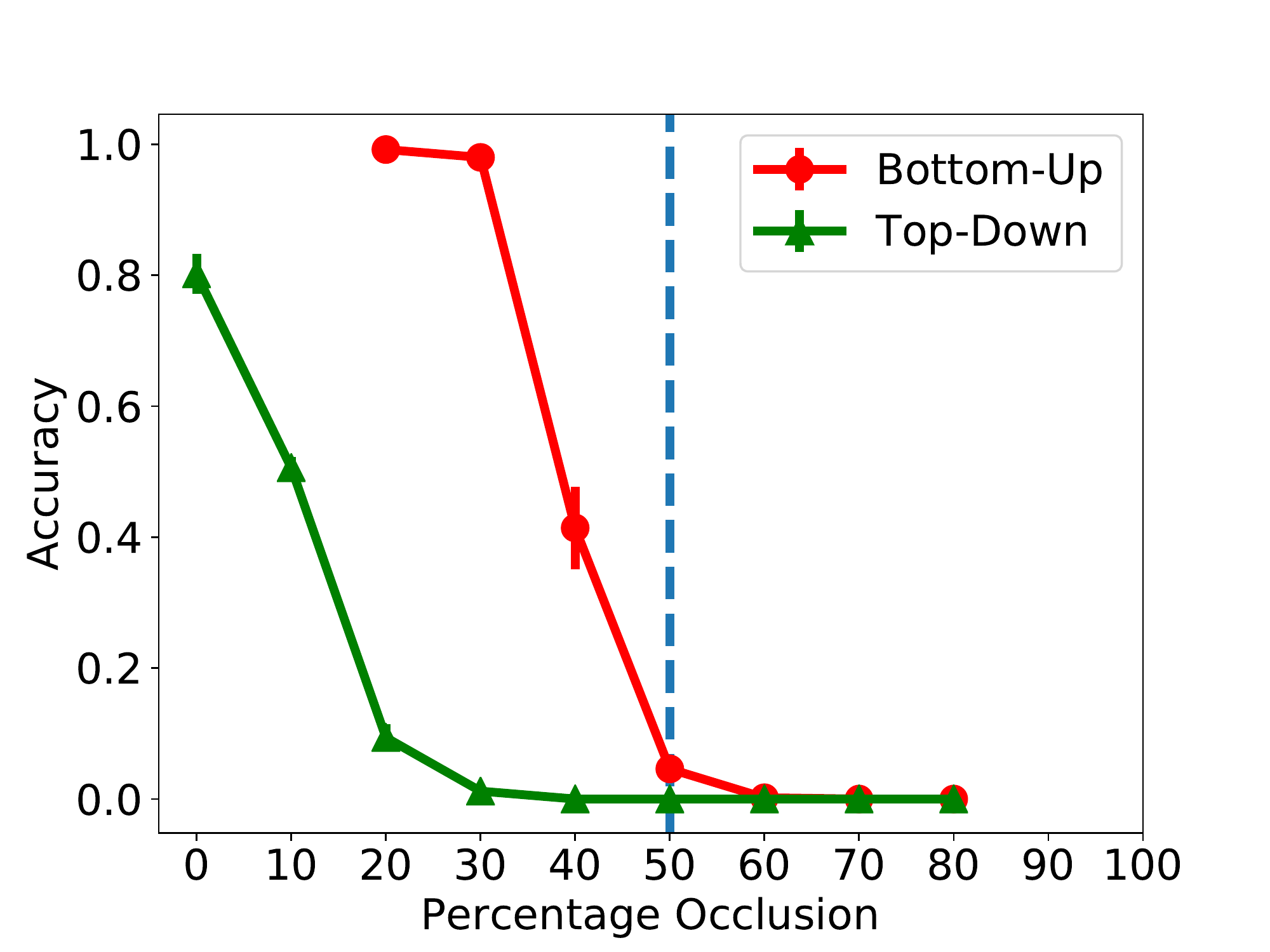} 
    \caption{Attn model (Seq2Seq)} 
    \label{btt_vs_ttb:a} 
    % \vspace{4ex}
  \end{subfigure}%% 
  \begin{subfigure}[b]{0.4\linewidth}
    \centering
    \includegraphics[width=1\linewidth]{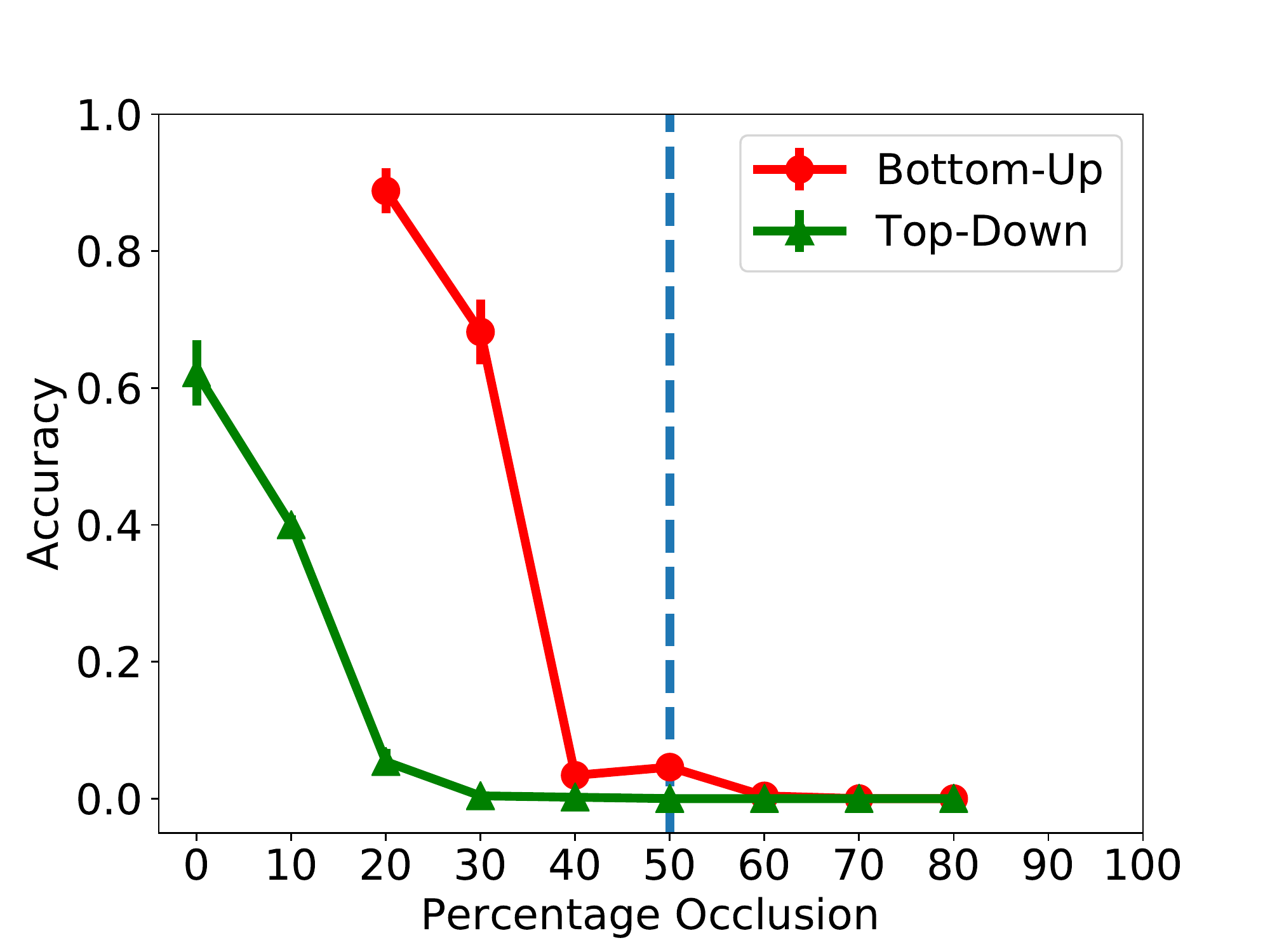} 
    \caption{CTC model (CRNN)} 
    \label{btt_vs_ttb:b} 
    % \vspace{4ex}
  \end{subfigure}
\caption{\emph{Comparing word accuracy performance of models on bottom-up occlusions vs top-down occlusions.} Models are trained on non-noisy images of all word lengths and evaluations are conducted on word images with different amounts of bottom-up (red) and top-down (green) occlusion. The blue dotted line in the middle indicates the point of 50\% occlusion. Average accuracy over 5 runs is plotted with error bars indicating +/- standard deviation.\vspace*{-0.4cm}}
  \label{btt_vs_ttb} 
\end{figure}

Like humans, both models perform better with top-down than with bottom-up occlusion. Fig.~\ref{btt_vs_ttb} shows that both models tolerate about 1.5 times bottom-up more than top-down occlusion. Sample occlusions are also shown in Fig.~\ref{percent_occlusion}. Both models tolerate much less occlusion than humans do \cite{huey1908psychology}.

\begin{figure*}[t!] 
\centering
  \begin{subfigure}[b]{0.4\linewidth}
    \centering
    \includegraphics[width=1\linewidth]{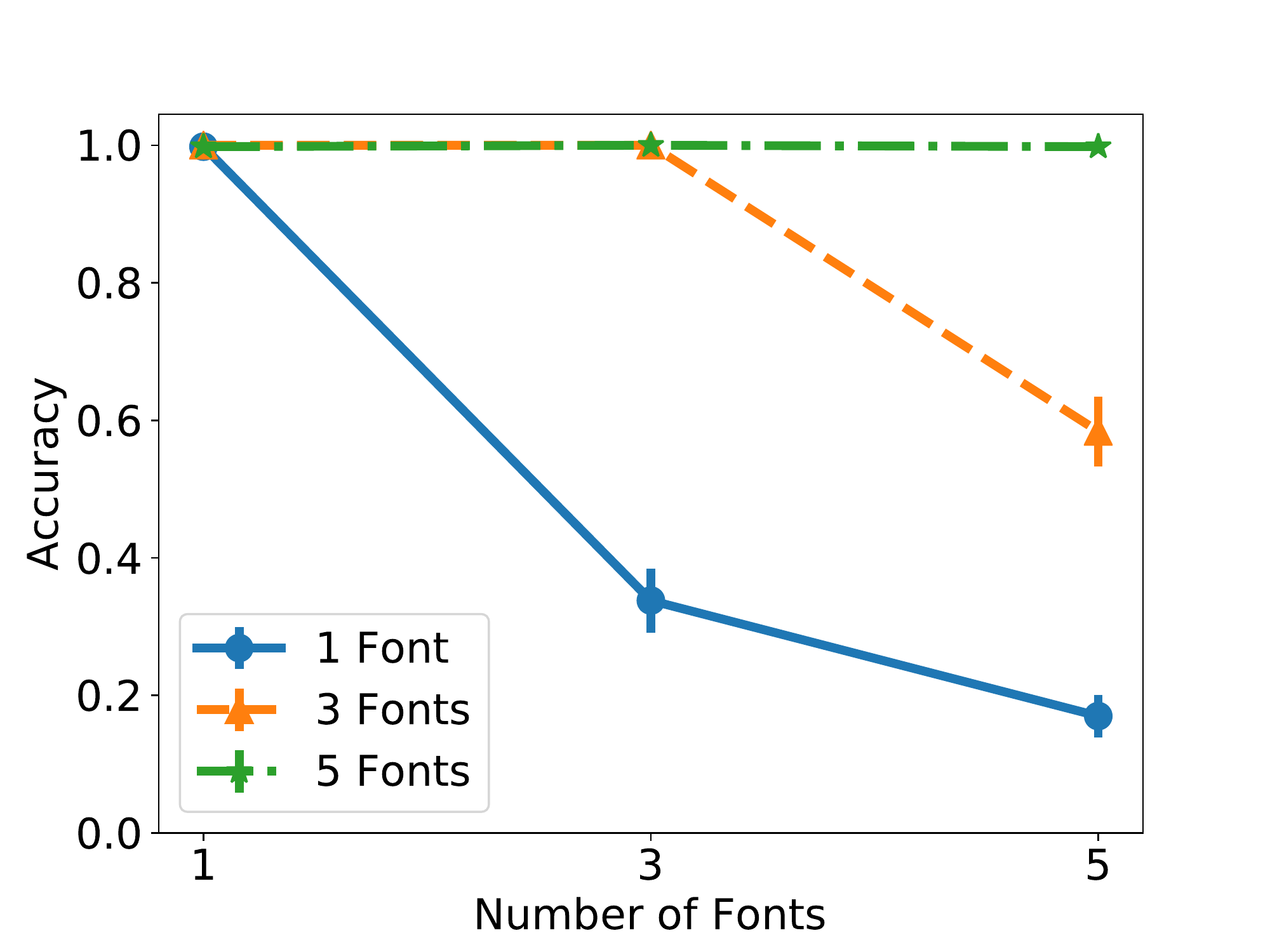} 
    \caption{Attn model (Seq2Seq)} 
    \label{font_compare:a} 
    % \vspace{4ex}
  \end{subfigure}%% 
  \begin{subfigure}[b]{0.4\linewidth}
    \centering
    \includegraphics[width=1\linewidth]{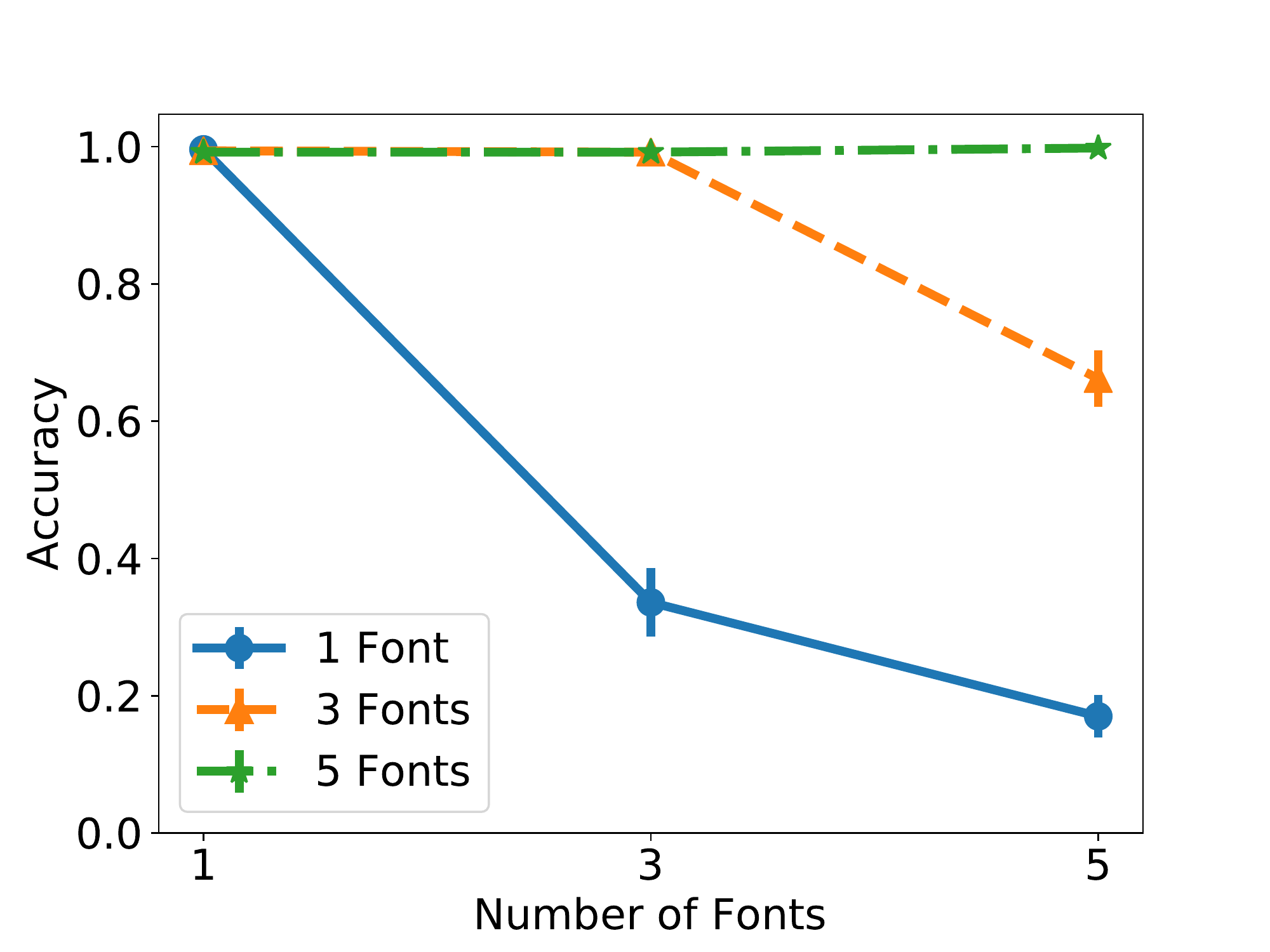} 
    \caption{CTC model (CRNN)} 
    \label{font_compare:b} 
    % \vspace{4ex}
  \end{subfigure}
\caption{\emph{Word accuracy performance on new fonts.} Models are trained and tested on noise-free images rendered with 1 font (blue), 3 fonts (orange), and 5 fonts (green). The five fonts used are - \protect\inlinegraphics{courier_img.pdf}, \protect\inlinegraphics{arial_img.pdf}, \protect\inlinegraphics{comicsansms_img.pdf}, {\protect\footnotesize\fontfamily{pbk}\selectfont Bookman Old Style}, and \protect\inlinegraphics{couriercttbolditalic_img.pdf}. Average accuracy over 5 runs is plotted as a function of number of fonts during testing. Error bars are +/- std. \vspace*{-0.6cm}}
  \label{font_compare} 
\end{figure*}

\noindent{\bf Fonts.} Unlike humans, both models fail to generalise to new fonts (Fig.~\ref{font_compare}), even when trained on multiple fonts. The CTC model performs marginally better than the Attn model when they are trained on 3 fonts and tested on 5 fonts. This suggests that the models rely heavily on the particular font shapes used at training, and are unable to generalize to slight changes in those letter shapes. 

\section{Conclusion}

Our results reveal the models' limited ability to generalize. The CTC model generalizes well to new word lengths while the Attn model does not. Both models tolerate less occlusion than humans, but they are like humans in tolerating more bottom-up than top-down occlusion. Training in noise improves both models' ability to generalize to occlusion. Lastly, both models generalize poorly to new fonts. Our experiments map the domain of generalization. These results demonstrate the value of testing models till they break, complementing the data science focus on optimizing performance. 

\clearpage
\section*{Broader Impact}
Neural networks today are widely used in diverse applications. Yet, those models usually suffer from the black-box problem, i.e. their failures are difficult to understand. For over a hundred years, human psychophysics has faced a similar challenge in characterizing performance of the human brain. 
In psychophysics, the brain is a black-box and its underlying mechanics are unveiled with behavioral experiments. One important contribution of psychophysics is the notion of a threshold. Complementing data science's emphasis on optimizing performance accuracy, psychophysical experiments push the system till it breaks. In that spirit, we measure the generalization threshold of a neural network model for word-length, occlusion, flankers, and number of fonts. Our experiments map the generalization domain of two scene text recognition models and reveal important limitations of these models. We recommend that the psychophysical practice of finding the breaking points be incorporated into testing of future models.

{\small
\bibliographystyle{plain}
\bibliography{bib_file}
}

\end{document}